\documentclass[letterpaper]{article}
\usepackage[preprint]{aaai2027}  % arXiv preprint version

\usepackage[hyphens]{url}
\usepackage{graphicx}
\usepackage{natbib}
\usepackage{caption}
\usepackage[utf8]{inputenc}
\usepackage{amsmath}
\usepackage{amsfonts}
\usepackage{amssymb}
\usepackage{booktabs}
\usepackage{makecell}
\usepackage{multirow}
\usepackage{pifont}
\usepackage{algorithm}
\usepackage{algorithmic}
\usepackage{colortbl}
\usepackage{array}

\usepackage{newfloat}
\usepackage{listings}
\DeclareCaptionStyle{ruled}{
    labelfont=normalfont,
    labelsep=colon,
    strut=off
}
\floatstyle{ruled}
\newfloat{listing}{tb}{lst}{}
\floatname{listing}{Listing}

\title{AptAvatar: Fast and Vivid Long-Form Audio-Driven Video Generation for Production-Ready Avatars}

\author{
    Hengyuan Zhang\equalcontrib,
    Jingna Sun\equalcontrib,
    Meiguang Jin\corresponding,
    Junfeng Ma
}

\affiliations{
    Alibaba Group, Taobao and Tmall Group
}

\begin{document}
\maketitle

% ===== Begin sec/0_abstract.tex =====
\begin{abstract}

%工业界即用方法需要->当前方法的妥协
Production-ready audio-driven avatar generation requires efficient inference without sacrificing fidelity or motion expressiveness. 
However, existing acceleration methods often compromise quality through restrictive architectural choices, such as causal attention and short temporal horizons, or by reducing model capacity and resolution.
% Production-ready audio-driven avatar video generation requires efficient and effective inference while preserving high-resolution visual fidelity and expressive motion dynamics; however, existing acceleration approaches often compromise these properties by enforcing unidirectional attention, shortening temporal horizons, or reducing model capacity, falling short of the demands of practical industrial deployment.
%介绍我们方法的特点
Without such compromises, we propose \textbf{AptAvatar}, a 14B-parameter
long-form audio-driven avatar generation framework that delivers fast and expressive inference.
% In this work, we propose \textbf{AptAvatar}, a 14B-parameter long-form audio-driven avatar generation framework that achieves fast and expressive inference without such compromises, thus guaranteeing generation quality.
% In this work, we propose \textbf{AptAvatar}, a high-quality two-step long-form audio-driven avatar video generation model that improves inference efficiency without resorting to these compromises.
%介绍问题难度
For efficiency in production-level applications, AptAvatar addresses the extreme two-step generation challenge.
% For efficiency in production-level applications, AptAvatar addresses the core difficulty of extreme two-step generation: aggressive step reduction opens a large endpoint-distribution gap, leaving the multi-step teacher too distant for direct optimization.
%创新点1
% To bridge this gap,
To bridge the gap between the multi-step teacher model and the two-step student model,
we introduce \textbf{Endpoint-Anchored Distribution
Distillation}. 
It augments vanilla distribution matching with a
dedicated \textbf{Anchor Score Estimator} trained on the trajectory-endpoint
distribution defined from a frozen pretrained 4-step bridge generator. 
This provides an attainable
endpoint-level anchor for the evolving two-step student.
%创新点2
To improve long-horizon consistency, we further introduce \textbf{Self-Generated
History Replay}, which reuses cached outputs from earlier generator checkpoints as
history conditions during chunk-wise training. 
This approximates inference-time conditioning on self-generated histories without costly online rollouts, 
% This approximates inference-time conditioning on self-generated histories without costly online rollouts, 
mitigating quality degradation from accumulated history errors.
%宣传我们的工作
Extensive experiments demonstrate that AptAvatar generates vivid
720p long-form avatar videos with only 2 NFEs, achieving a 60$\times$ speedup
while preserving visual fidelity and long-horizon identity.

\end{abstract}

\begin{figure*}[t]
  \centering
  \includegraphics[width=0.9\textwidth]{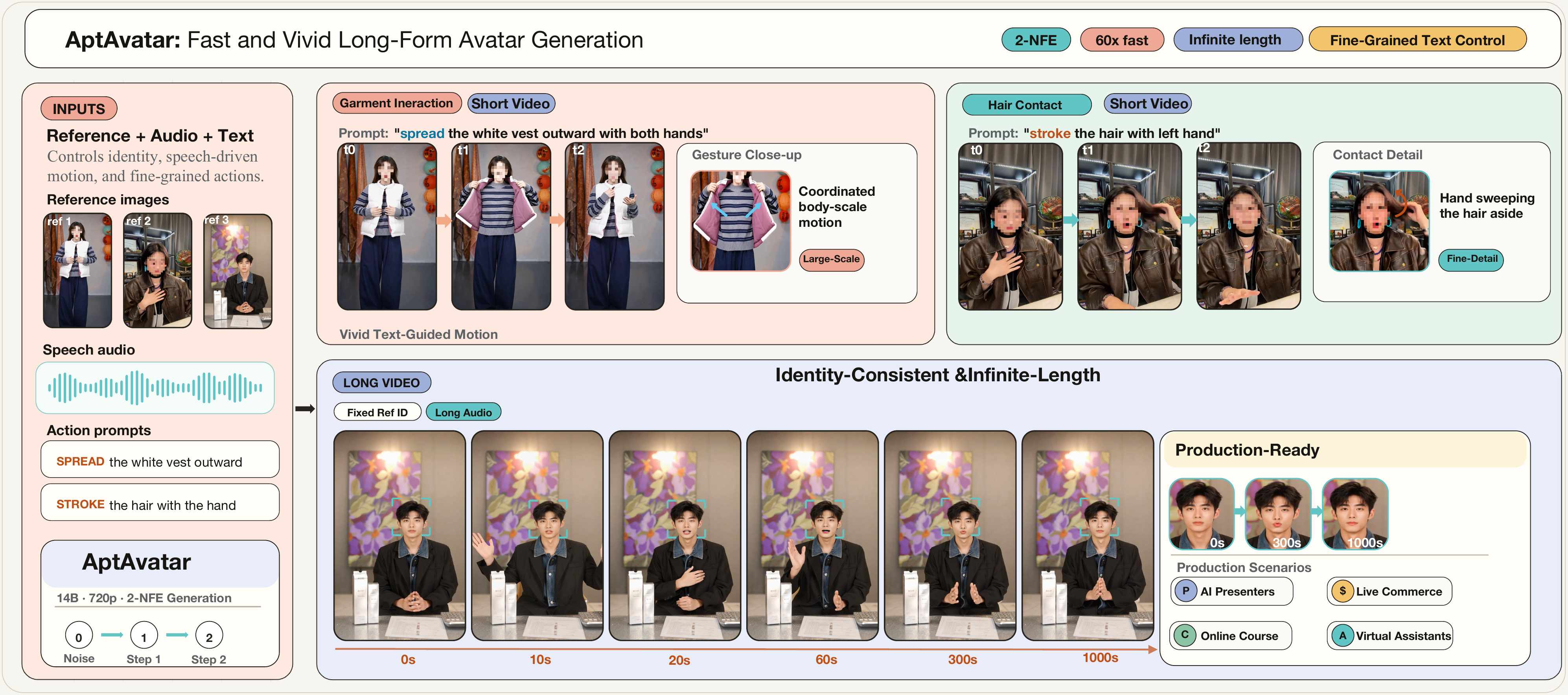}
  \caption{
    Teaser of \textbf{AptAvatar}. Given a reference identity, speech audio, and action prompts, AptAvatar generates 720p avatar videos with natural audio-driven motion and support for fine-grained text control over specific actions and interactions. 
    Using only $2$ NFEs, AptAvatar preserves vivid motion, identity consistency, and stable long-form synthesis, achieving a $60\times$ inference speedup for production-ready avatar applications.
  }
  \label{fig:teaser}
\end{figure*}
% ===== End sec/0_abstract.tex =====

% ===== Begin sec/1_intro.tex =====
\section{Introduction}
% Audio-driven avatar video generation aims to synthesize realistic human-centric videos from speech audio, reference images, and text prompts, and has become central to interactive digital humans.
We study audio-driven avatar video generation: given speech audio, a reference image, and a text prompt, the model synthesizes a realistic human-centric video with synchronized speech and temporally coherent motion.

Recent diffusion-based methods~\citep{gao2025wan,chen2025hunyuanvideo,team2026longcat} have markedly advanced the visual fidelity and motion expressiveness of audio-driven avatars. 
However, production-ready systems must additionally support efficient inference and stable long-form synthesis, creating a fundamental tension between generation quality and deployment efficiency. High-fidelity generation typically relies on large-capacity bidirectional models, long temporal windows, and multi-step denoising, whose computational cost grows rapidly with both spatial resolution and video duration. To improve efficiency, existing acceleration methods~\citep{yin2025slow,zhu2026causal,huang2026self} often introduce compromises such as causal attention, reduced model capacity, or shortened temporal windows, thereby weakening spatiotemporal modeling, motion expressiveness, and visual quality.

% %Production-ready avatar需求
% For production-ready avatars, the requirements go far beyond accurate lip synchronization: a practical system must deliver high-resolution fidelity, expressive facial and body motion, efficient inference, and stable long-form synthesis. 

% These goals are hard to satisfy jointly, as high-quality generation relies on large-capacity bidirectional models and long per-chunk temporal windows, whereas deployment demands efficient inference and robust long-horizon synthesis without sacrificing capacity, resolution, generation window, or bidirectional attention~\citep{huang2025live,shen2025soulx}.

% %当前模型进步很快
% Recent diffusion-based methods~\citep{gao2025wan,chen2025hunyuanvideo,team2026longcat} substantially improve visual fidelity and expressiveness, but their quality hinges on expensive multi-step denoising—especially prohibitive for high-resolution long-form generation, where spatial resolution and temporal duration jointly amplify computation. Existing acceleration methods~\citep{yin2025slow,zhu2026causal,huang2026self} alleviate this via architectural or computational compromises such as causal attention or reduced capacity, which improve speed but weaken bidirectional spatiotemporal reasoning, suppress motion expressiveness, and cap the visual quality required for production-ready avatars.

In this work, we target a more demanding setting: fast and vivid long-form audio-driven avatar generation without resorting to causal attention or shortened temporal windows. We pursue extreme efficiency via two-step denoising while preserving high-resolution fidelity and long-term consistency for production use. 
This setting poses two challenges: \textbf{(1) the endpoint-distribution gap} under extreme two-step distillation, and \textbf{(2) exposure bias} arising from the training--inference mismatch in long-form autoregressive generation.

For the first challenge, aggressive step reduction induces a large endpoint-distribution gap, leaving the multi-step teacher too distant to directly supervise two-step optimization. 
Existing methods bridge this gap with GAN-style discriminators~\citep{yin2024improved,liu2026high,feng2026one} or reference priors~\citep{wu2024freeinit,ren2024consisti2v}, yet the former is unstable and artifact-prone, while the latter biases the denoising trajectory and suppresses motion. 
For the second challenge, training conditions each chunk on clean history while inference relies on imperfect self-generated history, inducing exposure bias and cumulative identity drift. 
Recent methods~\citep{huang2026self,shen2025soulx} mitigate this mismatch via online self-rollouts but incur prohibitive compute and memory costs from repeatedly generating and propagating motion frames or KV caches.
To address these challenges, we propose \textbf{AptAvatar}, which distills a full-capacity 14B bidirectional diffusion model into a two-step generator while retaining 720p resolution and a 3-second temporal window per chunk, without causal attention or capacity reduction. This is enabled by two complementary designs.

%创新点1
First, we introduce Endpoint-Anchored Distribution Distillation (EADD) to stabilize extreme two-step distillation. Since early student samples can deviate substantially from the target distribution, the distant teacher score becomes ineffective for direct optimization. EADD augments standard distribution matching with an Anchor Score Estimator trained on the stationary endpoint distribution of a frozen pretrained four-step bridge generator. Unlike the fake score estimator, whose target drifts with the evolving student, the anchor score estimator provides a stable and attainable endpoint reference that complements the teacher score throughout two-step optimization. 
Second, we propose Self-Generated History Replay (SGHR) to mitigate long-horizon error accumulation without recurrent full-sequence rollouts. SGHR caches detached chunks generated by earlier student states and stochastically replays them as history conditions, exposing the model to diverse self-generated degradations while requiring only one new chunk per iteration. It therefore approximates inference-time self-conditioning at substantially lower training cost.

Extensive experiments demonstrate that AptAvatar generates high-quality 720p long-form avatar videos with only 2 NFEs, achieving a 60$\times$ inference speedup over the multi-step baseline~\citep{yang2025infinitetalk} while preserving visual fidelity, expressive motion, and long-horizon identity consistency.

Our main contributions are summarized as follows:
\begin{itemize}
% \item We present \textbf{AptAvatar}, a 14B-parameter framework for fast and vivid long-form audio-driven avatar generation.
% without causalizing the attention architecture or shortening the per-chunk temporal window.

% \item We introduce \textbf{Endpoint-Anchored Distribution Distillation}, which augments standard DMD with an anchor score estimator learned from the endpoint distribution of a frozen four-step bridge generator, providing a closer and more attainable reference for extreme two-step optimization.

% \item We propose \textbf{Self-Generated History Replay}, which stochastically replays cached chunks from earlier student states as history conditions, approximating inference-time self-conditioning while requiring only single-chunk generation at each training update.
\item \textbf{Stable Extreme Two-Step Distillation}:  We introduce Endpoint-Anchored Distribution Distillation for extreme two-step compression. It anchors standard DMD to the endpoint distribution of a frozen four-step bridge, enabling stable optimization while preserving visual fidelity.
% We introduce Endpoint-Anchored Distribution Distillation, which augments standard DMD with an anchor score estimator learned from the endpoint distribution of a frozen four-step bridge generator, providing a closer and more attainable reference for stable two-step optimization.

\item \textbf{Efficient Long-Horizon Self-Conditioning}: We propose Self-Generated History Replay to improve self-conditioning training. It replays cached student outputs as history conditions, mitigating quality degradation over long-horizon generation.
% We propose Self-Generated History Replay, which stochastically replays cached chunks from earlier student states as history conditions, approximating inference-time self-conditioning while requiring only single-chunk generation at each training iteration.
\item \textbf{Production-Ready Efficiency and Vividness}: 
AptAvatar achieves state-of-the-art performance in both short- and long-form 720p avatar generation. With only 2 NFEs, it delivers a 60$\times$ inference speedup while maintaining visual fidelity, expressive motion, and identity consistency.
% Extensive experiments demonstrate that \textbf{AptAvatar} achieves state-of-the-art performance in 720p short- and long-form avatar generation, attaining a 60$\times$ inference speedup with only 2 NFEs while preserving visual fidelity, expressive motion, and long-horizon identity consistency.

\end{itemize}

\begin{figure*}[t]
    \centering
    \includegraphics[width=0.95\linewidth]{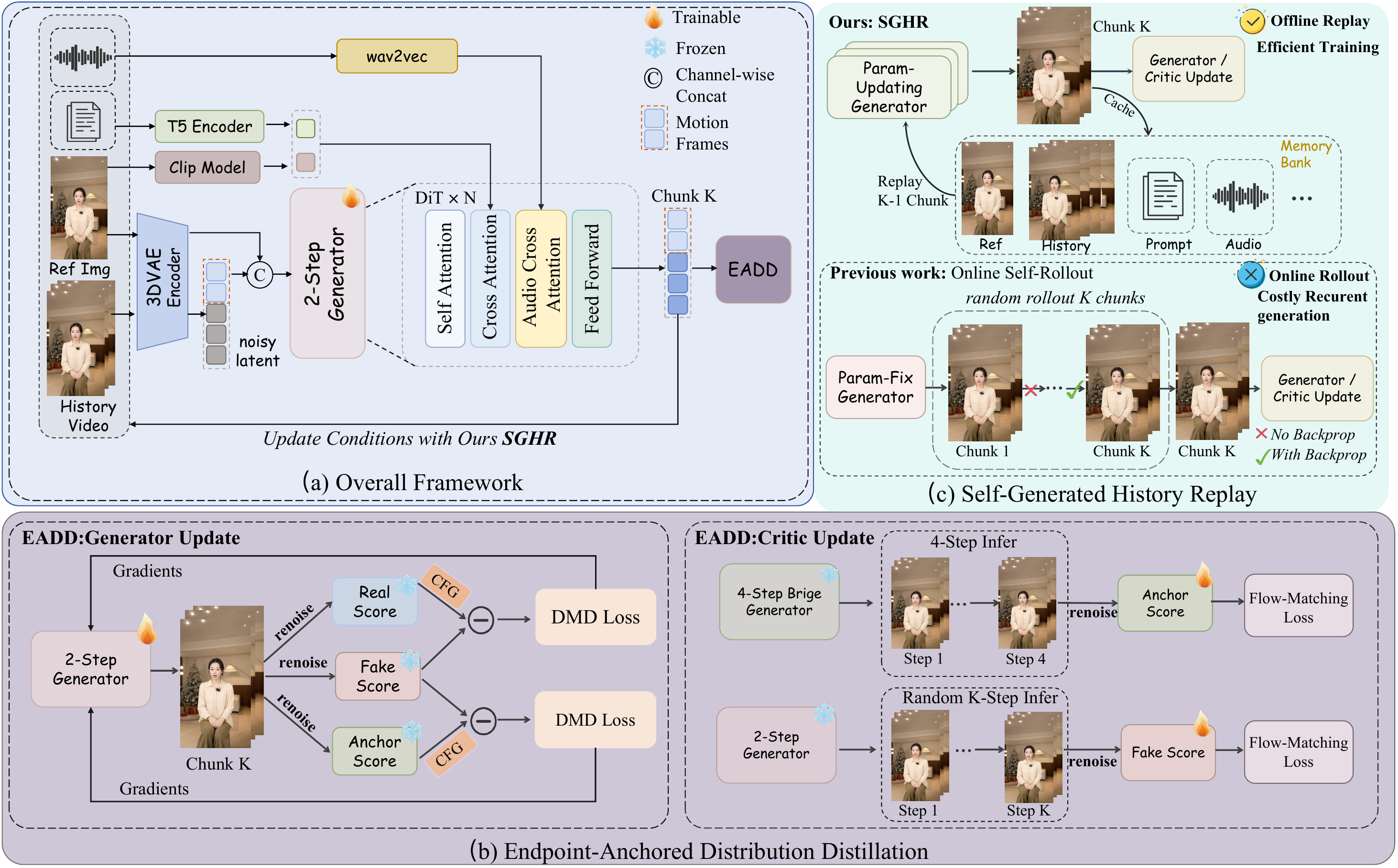}
        \caption{
        \textbf{Overview of AptAvatar.}
        (a) The overall chunk-wise framework encodes audio, text, reference image, and video history conditions, then generates the next chunk with a two-step DiT generator. EADD supervises two-step distillation, and SGHR updates the history conditions used during training.
        (b) EADD introduces an anchor score estimator learned from the endpoint distribution of a frozen four-step bridge generator, providing a stable target for four-to-two-step distillation.
        (c) SGHR replays cached self-generated chunks from earlier checkpoints as history conditions, approximating inference-time self-conditioning without costly online rollouts.
        }
    \label{fig:method_overview}
\end{figure*}
% ===== End sec/1_intro.tex =====

% ===== Begin sec/2_related_work.tex =====
\section{Related Work}

\subsection{Audio-driven Avatar Video Generation.}
% \paragraph{Audio-driven Avatar Video Generation.}
%任务是什么，需要什么
%发展阶段：diffusion前，diffusion后阶段，最近的基模阶段
%五句话！
%非端到端：wav2clip, sadtalk (使用3d flame， 3DMM), talking head or 半身 ; 基于gan的
%端到端：基于diffusion。基于dit, 比如wa-s2v, infinietalk, longcat-video-avatar 等
%多人和对话场景

Audio-driven avatar video generation aims to produce realistic talking videos from speech, text, and a reference image, with accurate lip synchronization, faithful identity preservation, and expressive motion.
Early methods decouple motion from appearance, relying on region-specific GAN synthesis~\citep{prajwal2020lip} or 3D motion representations with neural rendering~\citep{zhang2023sadtalker}, and mostly target facial or talking-head animation.
Diffusion-based methods instead cast the task as end-to-end conditional generation, jointly modeling appearance and audio-driven dynamics~\citep{tian2024emo,lin2025cyberhost}.
Recent video DiTs such as Wan-S2V~\citep{gao2025wan}, InfiniteTalk~\citep{yang2025infinitetalk}, and LongCat-Video-Avatar~\citep{team2026longcat} extend this to full-body, long-form, and cinematic generation.
MultiTalk~\citep{kong2026let} and HunyuanVideo-Avatar~\citep{chen2025hunyuanvideo} further support multi-person conversations by binding multiple speakers to their audio streams.

\subsection{Distribution Matching Distillation.}
% \paragraph{Distribution Matching Distillation.}

Diffusion distillation accelerates sampling via trajectory regression~\citep{frans2025one}, consistency learning~\citep{lu2025simplifying}, and distribution matching~\citep{yin2024one,yin2024improved}. 
Among these, Distribution Matching Distillation aligns the student's output distribution with a pretrained teacher at intermediate noise levels, enabling few-step generation while preserving its generative prior, and has since been strengthened with adversarial objectives~\citep{yin2024improved} and trajectory-aware formulations~\citep{luo2025learning}. 
Recent analyses further clarify its mechanism: Decoupled DMD~\citep{liu2025decoupled} attributes few-step conversion mainly to a CFG-augmentation engine with distribution matching as a stabilizing regularizer, while other work interprets DMD as reinforcement learning with the teacher as an implicit reward~\citep{jiang2025distribution}. 

\subsection{Long-Form Video Generation.}
% \paragraph{Long-Form Video Generation.}
Long-form video generation is commonly formulated as chunk-wise autoregression, where each segment conditions on bounded history represented by rolling KV caches, preceding motion frames, or compressed contexts~\citep{zhang2026frame}. 
Recent methods extend generation to minute-scale horizons through few-step causal distillation~\citep{yin2025slow}, autoregressive self-rollout
training~\citep{huang2026self}, and techniques such as KV recaching, frame-sink attention, and long-rollout supervision~\citep{yang2025longlive,cui2025self}. 
Similar paradigms have been adopted for streaming audio-driven avatars, including causal few-step models with long-horizon stabilization~\citep{huang2025live} and chunk-wise
bidirectional models with retrospective
correction~\citep{shen2025soulx}. 

\section{Method}
\label{sec:method}

% We present \textbf{AptAvatar}, a progressive distillation framework that compresses a multi-step audio-driven avatar diffusion model into an efficient two-step generator for long-form synthesis. 
% As illustrated in Fig.~\ref{fig:method_overview}, AptAvatar enables production-ready long-form avatar generation in only two denoising steps through two complementary innovations.
% First, \textbf{Endpoint-Anchored Distribution Distillation} trains an Anchor Score Estimator on the trajectory-endpoint distribution of a frozen four-step bridge generator. This endpoint distribution provides a more attainable distillation target, anchoring the two-step generator toward the real data distribution.
% Second, \textbf{Self-Generated History Replay} reuses cached outputs from previous checkpoints as history conditions, approximating inference-time self-conditioning without costly online rollouts and thereby improving long-horizon stability.
As illustrated in Fig.~\ref{fig:method_overview}, AptAvatar enables two-step long-form avatar generation through two complementary designs: \textbf{Endpoint-Anchored Distribution Distillation}, which stabilizes extreme two-step compression, and \textbf{Self-Generated History Replay}, which mitigates error accumulation in autoregressive long-form generation. 
We first describe the four-step bridge generator, then these two designs.

% As illustrated in Fig.~\ref{fig}, AptAvatar leverages a frozen four-step bridge generator to guide the training of a two-step generator with  two complementary designs.

% First, \textbf{Endpoint-Anchored Distribution Distillation} trains an Anchor Score Estimator on the trajectory-endpoint distribution of the frozen bridge generator, providing a more attainable target that anchors the two-step student generator toward the real distribution. 
% Second, \textbf{Self-Generated History Replay} reuses cached outputs from earlier checkpoints as history conditions, approximating inference-time self-conditioning without costly online rollouts and thereby improving long-horizon stability.

\subsection{Preliminaries: Four-Step Generator}
\label{sec:preliminaries}

We adopt InfiniteTalk~\citep{yang2025infinitetalk} as the backbone and fine-tune it on in-domain avatar data to obtain a strong bidirectional teacher, which is progressively distilled via ODE initialization and DMD into a four-step \emph{bridge generator} that provides an attainable target for subsequent two-step distillation.
% We adopt InfiniteTalk~\citep{yang2025infinitetalk} as the backbone and fine-tune a multi-step diffusion model on our in-domain avatar data, yielding a strong bidirectional teacher. To obtain a reliable intermediate model for subsequent two-step compression, we progressively distill this teacher into a four-step generator through two stages: (i) \emph{ODE initialization},  and (ii) \emph{distribution matching distillation} (DMD). We term the resulting four-step model the \emph{bridge generator}, which provides an attainable auxiliary target during distribution distillation.

\paragraph{ODE Initialization.} Following prior few-step video distillation~\citep{yin2025slow,huang2026self}, we use our fine-tuned teacher to generate deterministic ODE trajectories as supervision targets, and train a student $G_{\phi}$ to regress their endpoints. Denoting the clean sample from solving the ODE as $\mathcal{S}_{\psi}^{\mathrm{ODE}}(\cdot)$, the objective is: \begin{equation} \mathcal{L}_{\mathrm{ODE}}(\phi) = \mathbb{E}_{t,\,z_t} \left[ \left\| G_{\phi}(z_t,\mathbf{c}) - \mathcal{S}_{\psi}^{\mathrm{ODE}}(z_t,\mathbf{c}) \right\|_2^2 \right], \label{eq:pbd_ode_init} \end{equation} with $t$ from the discrete four-step schedule. This provides stable initialization for subsequent four-step distillation.

% \paragraph{ODE Initialization.} 
% Following prior work on few-step video distillation~\citep{yin2025slow,huang2026self,cui2025self,zhu2026causal}, we first employ our fine-tuned bidirectional multi-step teacher to generate deterministic ODE trajectories, which serve as supervision targets. We then train a bidirectional student generator $G_{\phi}$, initialized from the teacher weights, to regress the corresponding trajectory endpoints. Specifically, for each discrete timestep of the few-step scheduler, we denote the clean sample obtained by solving the ODE trajectory as $\mathcal{S}_{\psi}^{\mathrm{ODE}}(\cdot)$, and define the optimization objective as:
% \begin{equation}
% \mathcal{L}_{\mathrm{ODE}}(\phi) = \mathbb{E}_{t,\,z_t} \left[ \left\| G_{\phi}(z_t,\mathbf{c}) - \mathcal{S}_{\psi}^{\mathrm{ODE}}(z_t,\mathbf{c}) \right\|_2^2 \right],
% \label{eq:pbd_ode_init}
% \end{equation}
% where $t$ is sampled from the discrete four-step schedule. 
% This stage provides stable initialization for subsequent four-step distillation.

\paragraph{Four-Step Distribution Matching Distillation.}
Although the ODE-initialized generator typically exhibits low fidelity and blurred motion, it serves as a good starting point that accelerates
convergence and stabilizes the subsequent training. 
Building on it, we further apply Distribution Matching Distillation (DMD) to stably compress the sampling steps and distill the guided teacher into a
guidance-free student.

Specifically, DMD reduces the distributional divergence between the student and teacher at each noise level by minimizing the reverse Kullback--Leibler divergence, leading to the following tractable score-based gradient:
\begin{equation}
\nabla_{\phi}\mathcal{L}_{\mathrm{DMD}}
=
-\mathbb{E}_{t,\mathbf{z}}
\left[
\left(
s_{r}\!\left(\Psi(G_{\phi})\right)
-
s_{f}\!\left(\Psi(G_{\phi})\right)
\right)
\frac{\partial G_{\phi}}{\partial \phi}
\right],
\label{eq:dmd_gradient}
\end{equation}
where $s_{r}$ and $s_{f}$ denote the real and fake score estimators, given by the frozen teacher and the trainable critic, respectively.
$G_{\phi}$ denotes the four-step generator, and $\Psi$ represents the forward diffusion process. 

%是否要加个\paragraph{Anchor Score Estimator.}说明传统的gan和别的方法的对比说明
% 可以放在不同文章怎么桥接teacher和student，比如gan，噪声初始化。而我们提出了xxx，相比于他们，我们学习4步蒸馏的分布并利用这个分布给generator提供anchor 梯度，稳定整个extreme step reduction的训练，并达到工业级可用的状态 业务
%在哪强调这部分改进是为了快！同时，表述能够保持2-step的高质量标准，达到业界可用的状态
\subsection{Endpoint-Anchored Distribution Distillation}

Although the four-step model already provides substantial acceleration, further reducing the sampling budget to two steps is crucial for improving generation efficiency. 
However, direct four-to-two-step distillation suffers from severe quality degradation such as blurry motion and distorted hands,
since student-generated samples can drift far from regions well supported by the target distribution, so that the real score becomes unreliable and fails to provide an attainable optimization direction~\citep{bai2026optimizing, liu2026high, wu2026sgmd}.
To address this issue, we propose \textbf{Endpoint-Anchored Distribution Distillation} (EADD).

%现有两派方案
To mitigate this distribution gap, existing approaches typically introduce auxiliary guidance through either (i) a GAN-style discriminator that provides
adversarial supervision from real samples~\citep{yin2024improved,liu2026high,feng2026one,liu2026turbotalk,cheng2026phased}, or (ii) a reference prior injected into the initial noise to ease few-step denoising~\citep{wu2024freeinit,ren2024consisti2v, li2024tuning,yi2025magic}. 
%他们的缺点
However, the former is difficult to optimize because its coarse real--fake objective provides insufficient token-level guidance, often introducing artifacts, whereas the latter suppresses motion dynamics and reduces diversity.
%我们的创新（与他们不一样）
Therefore, distinct from these, we augment vanilla distribution matching with a dedicated \textbf{Anchor Score Estimator} trained on the trajectory-endpoint distribution defined by the frozen four-step bridge generator, which provides a more attainable and informative distribution to anchor the two-step generator toward the real distribution.

\paragraph{Endpoint Anchor Score.}
Let $G_{\theta}$ be the evolving two-step generator and $G_{\phi}$ the frozen
four-step bridge generator obtained before. Given noise $\mathbf{z}$ and condition $c$
(audio, reference image, text prompt, and motion frames), the bridge produces an endpoint
\begin{equation}
\hat{x}_{0}^{\phi}
=
G_{\phi}^{T\rightarrow 0}(\mathbf{z},\mathbf{c})
\label{eq:bridge_endpoint}
\end{equation}
through its full four-step inference trajectory. Here, $T \in \{4,2\}$ denotes the total number of denoising steps.
We then diffuse this endpoint
via the forward process $\Psi$ and train the anchor score estimator $s_{a}$
with a standard denoising objective:  %参考decoupled-dmd写
\begin{equation}
\mathcal{L}_{\mathrm{anchor}}
= 
\mathbb{E}_{t,\mathbf{z}}
\left[
\left\|
s_{a}\!\left(\Psi(G_{\phi})\right)
-
\hat{x}_0^{\phi}
\right\|_2^2
\right].
\label{eq:eadd_anchor_loss}
\end{equation}

% The key difference from the fake score estimator lies in the target distribution. 
Unlike the fake score estimator $s_f$ used in the current two-step distillation, the anchor score estimator is learned from a fixed target distribution. Specifically, $s_f$ must track a \emph{moving target}: the
2-step generator marginal at a randomly sampled exit timestep $t$, given by $\hat{z}_{t}^{\theta}=G_{\theta}^{T\rightarrow t}(\mathbf{z},\mathbf{c})$, which keeps shifting
as 2-step generator is updated. In contrast, the anchor score is fitted to the endpoint distribution of the \emph{frozen} bridge in
Eq.~\eqref{eq:bridge_endpoint}, which stays \emph{fixed} throughout training.
Distribution matching between the anchor and fake scores therefore guides the 2-step generator toward the bridge endpoint distribution, providing a closer and more attainable target than the distant distribution of real score.
% This time-invariant target offers the two-step generator a stable and readily attainable anchor rather than a constantly moving goal.
% Unlike the fake score estimator $s_f$ used in the current two-step distillation, the anchor score estimator is learned from a fixed target distribution. Specifically, $s_f$ tracks the evolving student marginal induced at a randomly sampled exit timestep $t$,
% $\hat{\mathbf{z}}{t}^{\theta}
% =G{\theta}^{T\rightarrow t}(\mathbf{z},\mathbf{c})$,
% which continuously shifts as the two-step generator is updated. In contrast, $s_a$ models the endpoint distribution of the frozen four-step bridge defined in Eq.~\eqref{eq}, and thus remains fixed throughout training. Distribution matching between the anchor and fake scores therefore guides the student toward the bridge endpoint distribution, providing a closer and more attainable target than the distant real distribution and stabilizing extreme two-step optimization.

\paragraph{Anchor-Guided Generator Update.}

Following DMD, we update the two-step generator  using the score
difference between the anchor estimator and the fake estimator:
\begin{equation}
\nabla_{\theta}\mathcal{L}_{\mathrm{EADD}}
=
-\mathbb{E}_{t,\mathbf{z}}
\left[
\left(
s_{a}\!\left(\Psi(G_{\theta})\right)
-
s_{f}\!\left(\Psi(G_{\theta})\right)
\right)
\frac{\partial G_{\theta}}{\partial \theta}
\right].
\label{eq:eadd_generator_grad}
\end{equation}
Unlike the real score, which attracts the generator toward the distant
teacher distribution, the anchor score provides a closer and more attainable
endpoint-level target, thereby offering more reliable guidance for two-step
optimization.

We combine this endpoint-anchored gradient with the standard DMD gradient:
\begin{equation}
\nabla_{\theta}\mathcal{L}_{\mathrm{gen}}
=
\nabla_{\theta}\mathcal{L}_{\mathrm{DMD}}
+
\lambda_{\mathrm{eadd}}\,\nabla_{\theta}\mathcal{L}_{\mathrm{EADD}},
\label{eq:eadd_total_generator}
\end{equation}
where the DMD term promotes alignment with the teacher through the real score, while
the EADD term guides the 2-step generator toward the more accessible bridge-endpoint
distribution through the anchor score. Their combination reduces the difficulty of
directly matching a distant teacher while retaining teacher-aligned
supervision. 

Since the anchor serves as a proxy for the real score, we generate the bridge endpoints in Eq.~\eqref{eq:bridge_endpoint} under random condition dropout, aligning its behavior with the classifier-free guidance of the real score and thereby stabilizing the 2-step generator's conditional training.

%%创新点3
%注意是否要补充，我们的offline可以比online效果更好！！！提升了多少倍速度！！！
%是否要补充流程图！！！
\subsection{Self-Generated History Replay}
\paragraph{Rethinking online-rollout history alignment.}
In chunk-wise autoregressive video synthesis, conditioning on \emph{ground-truth} history during training but on \emph{self-generated} history during inference creates a history-conditioning exposure bias that accumulates errors over long horizons.

Existing methods~\citep{cui2025self,shen2025soulx} typically rely on recurrent online self-rollouts during training, propagating history through generated motion frames or KV caches and thereby incurring substantial compute and memory overhead, especially in our 720p setting without shortening the temporal horizon.

We instead argue that faithful history alignment need \emph{not} regenerate the
current model's exact history online. What matters is exposing the generator to a
sufficiently representative distribution of \emph{self-generated} degradations
encountered at inference. Motivated by replay buffers in reinforcement learning ~\citep{lin1992self,mnih2015human},
we introduce a \textbf{Memory Bank} that caches chunks from earlier
generator checkpoints and reuses them as history conditions. These cached chunks
serve a dual role: they approximate online rollouts without regenerating them,
and the degraded outputs accumulated across past generator states further enrich
the history distribution seen by the generator.

\paragraph{Stack-Based Replay with Stochastic Depth.}
We maintain a stack-structured memory bank $\mathcal{B}$, whose each entry
$(\hat{\mathbf{x}}_0^{\theta},\,\mathbf{m},\,K)$ stores a detached clean chunk
$\hat{\mathbf{x}}_0^{\theta}$, its memory context
$\mathbf{m}=\{\mathbf{c}_{\mathrm{txt}},\mathbf{c}_{\mathrm{ref}},\mathbf{h}\}$
---comprising the text prompt, reference image, and history motion frames
---and the replay depth $K$. Together with the driving audio
$\mathbf{c}_{\mathrm{aud}}$, it forms the complete condition
$\mathbf{c}=\mathbf{m}\cup\{\mathbf{c}_{\mathrm{aud}}\}$. 
In both generator- and
critic-update phases, the generator produces only a \emph{single}
chunk per iteration, with its context memory context randomly sampled via $p\sim\mathcal{U}(0,1)$ from either the stack top or the ground-truth context.
\begin{equation}
(\mathbf{m},K)=
\begin{cases}
\operatorname{top}(\mathcal{B}), & p\le p_{\mathrm{rep}}\ \text{and}\ \mathcal{B}\neq\varnothing,\\[3pt]
(\mathbf{m}^{\mathrm{gt}},\,0), & \text{otherwise},
\end{cases}
\label{eq:sghr_router}
\end{equation}
Conditioned on this context, the generator unrolls the $(K{+}1)$-th chunk from
pure noise. For the training gradient, a timestep is randomly
sampled from the few-step schedule , and the loss is evaluated
on the intermediate output $G_{\theta}^{(K+1,\,t)}(\mathbf{z},\mathbf{c})$, the
result of denoising this chunk from pure noise down to timestep $t$. To
refresh the memory bank, the generator then continues the denoising to
$t\!=\!0$, and the resulting clean chunk is detached and pushed back with
incremented depth:
\begin{equation}
\mathcal{B}\leftarrow\operatorname{Push}\bigl(\mathcal{B},
(\operatorname{sg}(G_{\theta}^{(K+1,\,0)}),\,\mathbf{m}',\,K{+}1)\bigr).
\label{eq:sghr_push}
\end{equation}
where $\operatorname{sg}(\cdot)$ is the stop-gradient, $\mathbf{m}'$ denotes the
memory context updated with $G_{\theta}^{(K+1,\,0)}$ as the new history, and
the cached memory chain is reset once $K\!=\!K_{\max}$. 
Conditioning on the stack top thus mimics the $(K{+}1)$-th chunk of an online rollout, yet keeps gradients within the current chunk alone.

Under this scheme, the generator is updated by
\begin{equation}
\nabla_{\theta}\mathcal{L}
=\mathbb{E}_{\;t\sim\mathcal{T}^{(k)}, \mathbf{z}, K}
\Bigl[\nabla_{\theta}\,\mathcal{L}_{\mathrm{gen}}
\bigl(G_{\theta}^{(K+1,\,t)}(\mathbf{z},\mathbf{c})\bigr)\Bigr],
\label{eq:sghr_grad}
\end{equation}
and the fake-score and anchor score estimators are likewise trained on the same chunk $G_{\theta}^{(K+1,\,t)}$. 

% Since $\theta$ keeps evolving, chunks within a chain stem from different generator
% states, naturally covering heterogeneous history degradations.
Since $\theta$ continuously evolves during training, chunks within the same replay chain are generated by different historical generator states, naturally exhibiting heterogeneous levels and patterns of degradation. 
Replaying such mixed-quality histories exposes the current generator to a broader range of imperfect contexts, rather than a fixed error distribution. This improves its robustness to accumulated self-generated errors during long-form inference.

\section{Experiments}

\subsection{Experimental Settings}

\begin{figure*}[t]
    \centering
    \includegraphics[width=0.9\textwidth]{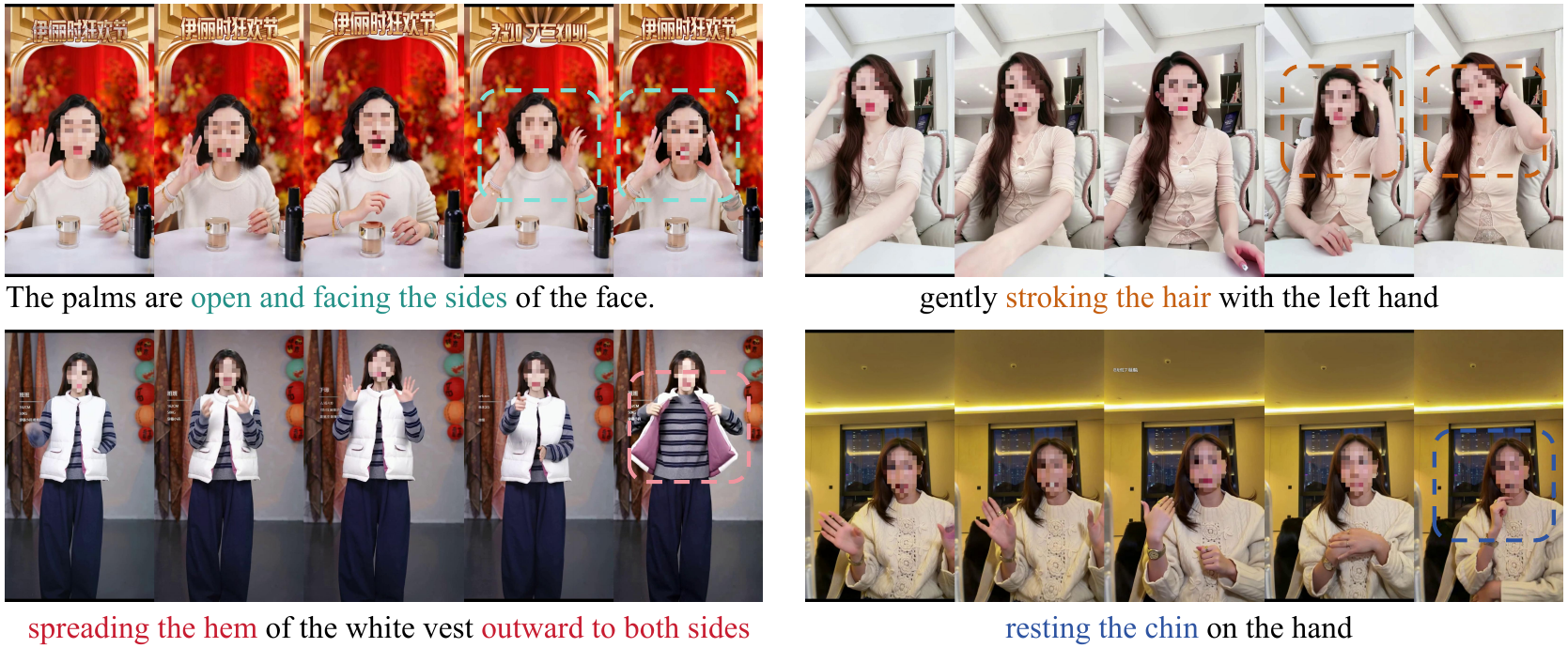}
    \caption{
        Fine-grained text-guided motion generation. AptAvatar faithfully follows action-specific textual prompts and produces the corresponding expressive movements.
    }
    \label{fig:text}
\end{figure*}

\textbf{Implementation. } We adopt InfiniteTalk~\citep{yang2025infinitetalk} as the backbone architecture and optimize it for the AptAvatar framwork. The training pipeline consists of two stages. First, supervised fine-tuning (SFT) stage is performed to adapt to the current data distribution and incorporate historical frames. Second,  based on the Self-Forcing training framework~\citep{huang2026self}, we  build our AptAvatar framwork. The training learning rates are set to $2 \times 10^{-6}$ for the generator and $4 \times 10^{-7}$ for the fake score model and the Anchor Score Estimator, with a 1:5 update ratio. Experiments are carried out on 32 NVIDIA H20 GPUs, using a batch size of 1 per GPU.

\begin{figure*}[!t]
    \centering
    \includegraphics[width=0.9\textwidth]{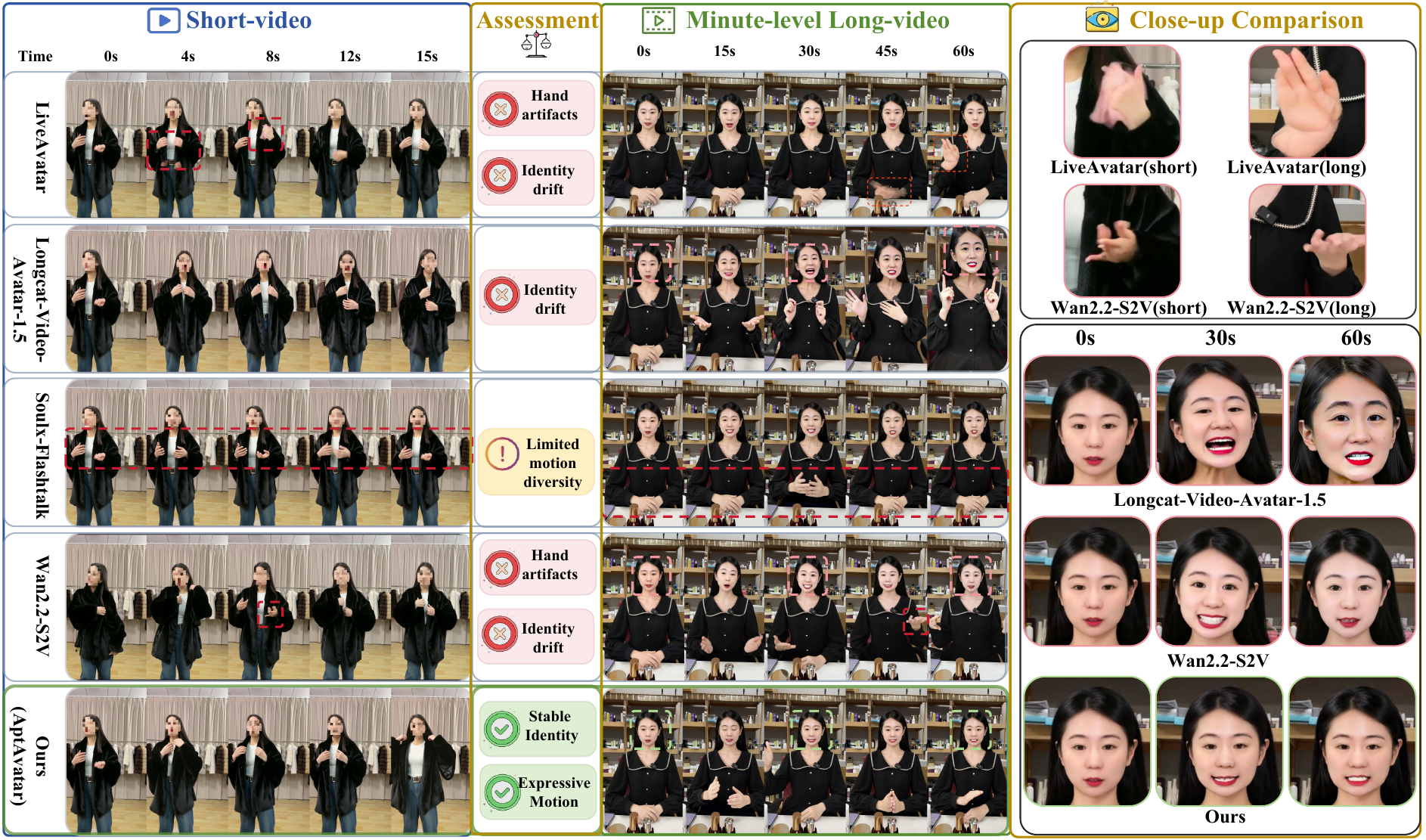}
    \caption{Visual comparison on short-video and minute-level long-video test sets. AptAvatar preserves hand integrity, identity consistency, and expressive motion across both short- and long-form generation.}
    \label{fig:visual-comparison}
\end{figure*}

\noindent
\textbf{Datasets.} We collect a large-scale video dataset of approximately 400 hours, which is used consistently across all training stages. The dataset covers diverse scenarios, including seated and standing poses, as well as various background settings. To better evaluate the model, we construct two test datasets: one contains 100 videos of 15 seconds each, with faces mosaicked due to copyright issues, and the other contains 20 videos of 1 minute each.

\noindent
\textbf{Evaluation Metrics.} For evaluation, we adopt a comprehensive set of metrics to assess both visual quality and temporal consistency. Specifically, we use the Aesthetics Score Evaluation (ASE)~\citep{wu2023q} to evaluate aesthetic quality. Lip-audio synchronization is measured by Sync-C and Sync-D~\citep{chung2016out}. We further employ VBench~\citep{huang2024vbench} to evaluate temporal quality from multiple perspectives, including Subject Consistency (Subject-C), Background Consistency (BG-C), Motion Smoothness (Motion-S), and Temporal Flickering (Temporal-F). Additionally, we report Fréchet Inception Distance (FID) for frame-level visual fidelity.

\subsection{Main Result}

\begin{table*}[!t]
    \centering
    \small
    \setlength{\tabcolsep}{2.2pt}
    \begin{tabular}{@{}lccccccccc@{}}
    \toprule
         Methods & NFE & FID $\downarrow$ & Subject-C $\uparrow$ & BG-C $\uparrow$ & Motion-S $\uparrow$ & Temporal-F $\uparrow$ & ASE $\uparrow$ & Sync-C $\uparrow$ & Sync-D $\downarrow$\\
    \midrule
    \rowcolor{gray!12}
    % \multicolumn{10}{c}{\textit{Short-video benchmark: 100 videos, 15 seconds each}}\\
    \multicolumn{10}{c}{\textit{Short-video benchmark}}\\
    \midrule
           InfiniteTalk & 120 & 24.808 & 96.889 & 95.568 & 99.512 & 99.318 & 3.146 & 7.172 & 8.010\\
           Wan2.2-S2V & 80 & 43.542 & 97.151 & 95.663 & 99.385 & 99.216 & 3.119 & 5.626 & 9.376\\
           LongCat-Video-Avatar 1.5 & 8 & 36.582 & 96.338 & 94.533 & 99.550 & 99.332 & 3.191 & 4.140 & 11.433\\
           LiveAvatar & 4 & 37.868 & 97.804 & 95.718 & 99.409 & 99.226 & 3.184 & 5.986 & 8.586\\
           SoulX-FlashTalk & 4 & 28.039 & 98.580 & 96.530 & 99.550 & 99.434 & 3.172 & 7.743 & 7.912\\
    \rowcolor{gray!6}
           \textbf{AptAvatar (Ours)} & \textbf{2} & \textbf{24.759} & \textbf{98.589} & \textbf{96.844} & \textbf{99.561} & \textbf{99.456} & \textbf{3.307} & \textbf{7.865} & \textbf{7.739}\\
    \midrule
    \rowcolor{gray!12}
    % \multicolumn{10}{c}{\textit{Minute-level long-video benchmark: 20 videos, 1 minute each}}\\
    \multicolumn{10}{c}{\textit{Minute-level long-video benchmark}}\\
    \midrule
           InfiniteTalk & 120 & -- & 96.670 & 94.897 & 99.579 & 99.371 & 3.149 & 8.178 & 7.870\\
           Wan2.2-S2V & 80 & -- & 97.037 & 94.883 & 99.491 & 99.402 & 3.186 & 6.173 & 9.541\\
           LongCat-Video-Avatar 1.5 & 8 & -- & 95.374 & 93.111 & 99.559 & 99.373 & 3.274 & 7.492 & 8.854\\
           LiveAvatar & 4 & -- & 98.115 & 96.001 & 99.452 & 99.310 & 3.272 & 6.654 & 9.008\\
           SoulX-FlashTalk & 4 & -- & 98.840 & 96.306 & 99.601 & 99.533 & 3.246 & 7.976 & 8.316\\
    \rowcolor{gray!6}
           \textbf{AptAvatar (Ours)} & \textbf{2} & -- & \textbf{98.895} & \textbf{96.489} & \textbf{99.634} & \textbf{99.623} & \textbf{3.382} & \textbf{8.398} & \textbf{7.775}\\
    \bottomrule
    \end{tabular}
    \caption{Quantitative comparison with state-of-the-art audio-driven avatar generation methods on short-video and minute-level long-video benchmarks. Best results are in bold.}
    \label{tab:sota}
\end{table*}

\begin{table*}[!t]
    \centering
    \small
    \setlength{\tabcolsep}{2.2pt}
    \begin{tabular}{@{}lccccccccc@{}}
    \toprule
         Variant & NFE & FID $\downarrow$ & Subject-C $\uparrow$ & BG-C $\uparrow$ & Motion-S $\uparrow$ & Temporal-F $\uparrow$ & ASE $\uparrow$ & Sync-C $\uparrow$ & Sync-D $\downarrow$\\
    \midrule
    \rowcolor{gray!12}
    \multicolumn{10}{c}{\textit{Short-video benchmark}}\\
    \midrule
           Vanilla DMD & 4 & 27.825 & 98.389 & 96.717 & 99.529 & 99.393 & 3.178 & 7.685 & 7.754\\
           Ours w/o ASE & 2 & 29.212 & 98.552 & 96.819 & 99.537 & 99.424 & 3.162 & 7.611 & 7.749\\
           \rowcolor{gray!6}
           \textbf{Ours} & \textbf{2} & \textbf{24.759} & \textbf{98.589} & \textbf{96.844} & \textbf{99.561} & \textbf{99.456} & \textbf{3.307} & \textbf{7.865} & \textbf{7.739}\\
    \midrule
    \rowcolor{gray!12}
    \multicolumn{10}{c}{\textit{Minute-level long-video benchmark}}\\
    \midrule

            Vanilla DMD w/ SGHR & 4 & -- & 98.472 & 96.202 & 99.611 & 99.559 & 3.241 & 8.250 & 7.834\\
            Vanilla DMD w/ online-rollout  & 4 & -- & 98.390 & 96.198 & 99.606 & 99.476 & 3.207 & 8.195 & 7.906\\
            Ours w/o ASE & 2 & -- & 98.882 & 96.472 & 99.632 & 99.615 & 3.237 & 8.322 & 7.807\\
           \rowcolor{gray!6}
           \textbf{Ours} & \textbf{2} & -- & \textbf{98.895} & \textbf{96.489} & \textbf{99.634} & \textbf{99.623} & \textbf{3.382} & \textbf{8.398} & \textbf{7.775}\\
    \bottomrule
    \end{tabular}
    \caption{Ablation study of AptAvatar on short-video and minute-level long-video benchmarks. ``w/o ASE'' removes the Anchor Score Estimator, using vanilla DMD. The online-rollout row is kept for the corresponding history-alignment variant.}
    \label{tab:ablation}
\end{table*}

\textbf{Quantitative  Analysis.}  We compare AptAvatar with recent state-of-the-art audio-driven avatar methods. For multi-step diffusion models,  InfiniteTalk~\citep{yang2025infinitetalk} and Wan-S2V~\citep{gao2025wan} are the typical audio-driven methods and provide a benchmark of metrics under multi-step models. For few-step long video generator models, the LiveAvatar~\citep{huang2025live}, SoulX-FlashTalk~\citep{shen2025soulx}, and LongCat-Video-Avatar~\citep{team2026longcat} are the recent attention-grabbing long-video generation approaches. They can extend video generation to the minute scale, which is particularly challenging for long-horizon generation.

% Table\ref{tab:qc} shows the quantitative metrics compared with these methods. 'ours-w/o ASE' represents the model is trained without Anchor Score Estimator, which is a plain 2 NFE DMD. The results demonstrate that under the 2-NFE limitation, our method achieves the best performance on FID, Subject-C, BG-C , Temporal-F, Sync-C and achieves the second-best performance on the other metrics on Motion-S and Sync-D.  The Motion-S metric assigns higher scores to smoother motions. The high score achieved by SoulX-FlashTalk can be attributed to its relatively conservative motion generation, characterized by lower motion diversity and smaller motion amplitudes.  The LongCat-Video-Avatar achieves the highest score on the ASE metric, which can be attributed to its smoother and more gentle motions. However, its main limitation in long-video generation lies in the degradation of visual quality and identity consistency over time, which is shown in Fig.~\ref{fig:visual-comparison}.

Table~\ref{tab:sota} reports the quantitative comparison with recent state-of-the-art methods on both short-video and minute-level long-video benchmarks. 
With only 2 NFEs,  AptAvatar achieves the best result on every reported metric across both benchmarks, demonstrating significant advantages in visual quality, temporal consistence, and audio--visual synchronization.
SoulX-FlashTalk ranks second on most metrics, yet still falls notably short of AptAvatar in short-video FID and ASE. Moreover, these aggregate metrics alone do not capture all aspects of long-video generation.  As shown in Fig.~\ref{fig:visual-comparison}, these methods suffer from visible identity drift, hand artifacts, or limited motion when extended to long-video generation.

\noindent
\textbf{Qualitative Analysis.} Fig.~\ref{fig:visual-comparison} presents sampled frames from the two test sets. LiveAvatar suffers from incomplete hands, SoulX-FlashTalk produces limited motion, and LongCat-Video-Avatar and Wan-S2V degrade in identity consistency and visual quality on minute-level videos. In Fig.~\ref{fig:text}, our method also responds more faithfully to fine-grained text instructions, such as ``spreading the hem of the white vest outward to both sides,'' while competing methods often miss the intended motion. Overall, AptAvatar better preserves identity and visual quality while generating more diverse and text-responsive motions.

\begin{figure}[!t]
    \centering
    \includegraphics[width=0.9\columnwidth]{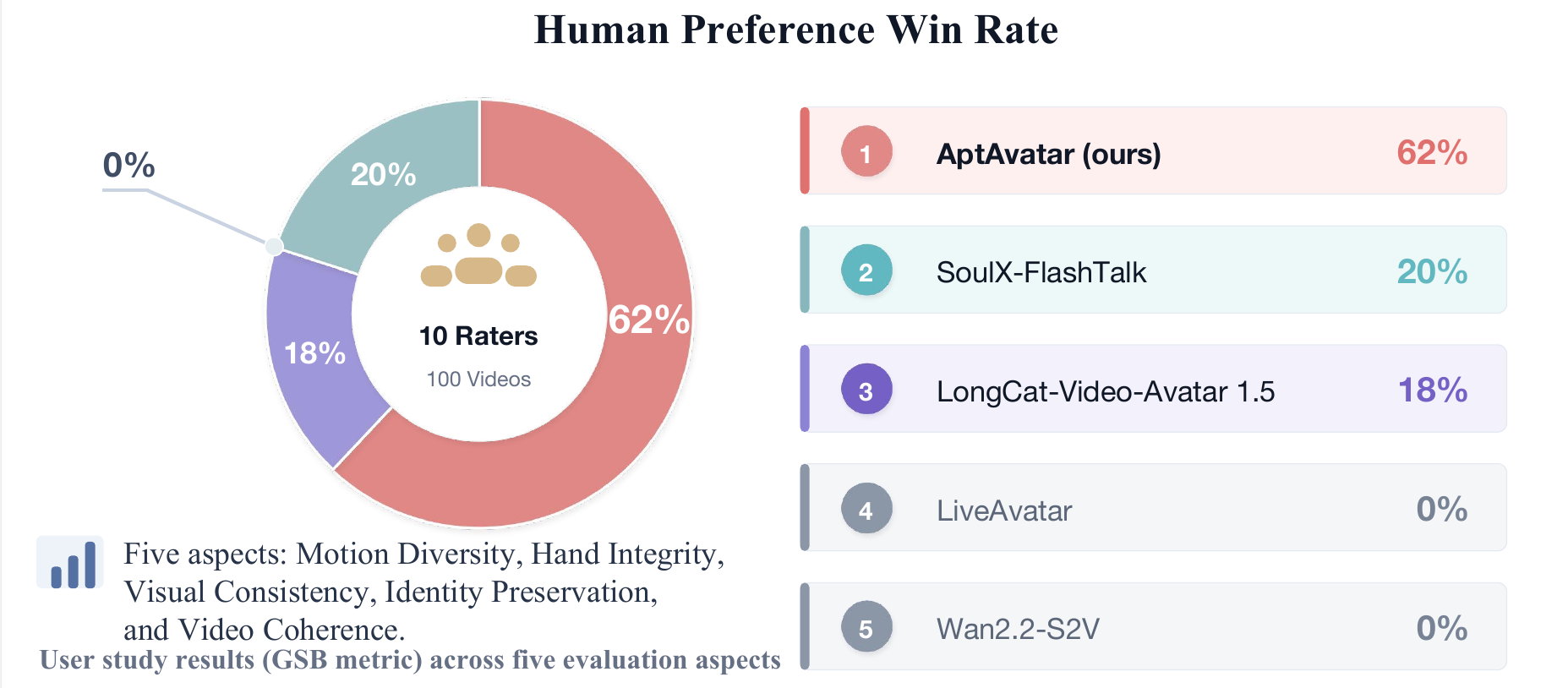}
    \caption{Human preference win rate under the GSB protocol. }
    \label{fig:user-study-win-ratio}
\end{figure}

% \textbf{User study.} We evaluate all methods via a user study using the GSB (good-same-bad) metric. We evaluate the generated videos from five aspects: motion diversity, hand integrity, visual consistency, identity preservation, and video coherence. The evaluation is performed on our collected dataset, which contains 100 videos of 15 seconds each. We recruit 10 participants as human raters. For a fair comparison, videos generated by different methods are randomly shuffled and anonymized before being presented to the raters. As shown in Fig.~\ref{fig:user-study-win-ratio}, AptAvatar obtains the highest win ratio, with 62\% of the votes, compared with 0\% for LiveAvatar, 20\% for SoulX-FlashTalk, 18\% for LongCat-Video-Avatar 1.5, and 0\% for Wan2.2-S2V. It is worth noting that our method obtains the most votes despite requiring only 2 NFEs, clearly demonstrating its effectiveness.
\noindent
\textbf{User Study.} We further conduct a GSB user study on 100 short videos, evaluating motion diversity, hand integrity, visual consistency, identity preservation, and video coherence. Videos from different methods are randomly shuffled and anonymized for 10 raters. As shown in Fig.~\ref{fig:user-study-win-ratio}, AptAvatar obtains the highest win ratio, with 62\% of the votes, compared with 0\% for LiveAvatar, 20\% for SoulX-FlashTalk, 18\% for LongCat-Video-Avatar, and 0\% for Wan2.2-S2V, demonstrating strong perceptual quality under only 2 NFEs.

\subsection{Ablation Study}
Table~\ref{tab:ablation} validates the contribution of each component. Compared with the 4-step DMD baseline introduced in the preliminaries, the full model achieves stronger overall quality and consistency despite using only 2 NFEs. Removing the Anchor Score Estimator degrades FID, ASE, Sync-C, and several consistency metrics, showing that endpoint anchoring provides more reliable guidance for extreme two-step distillation. 
The online-rollout variant further compares history-alignment strategies. Our SGHR-based design achieves strong long-video consistency without recurrent online rollouts, indicating that replaying cached self-generated histories is an effective and efficient substitute for expensive online history simulation.
% ===== End sec/4_experiments.tex =====

% ===== Begin sec/5_conclusion.tex =====
\section{Conclusion}
We presented \textbf{AptAvatar}, a novel framework for fast and vivid long-form audio-driven avatar generation. By combining Endpoint-Anchored Distribution Distillation with Self-Generated History Replay, AptAvatar transforms a powerful multi-step generator into an efficient deployment-ready system  requiring only 2 NFEs. On 720p short- and long-form benchmarks, AptAvatar attains a 60$\times$ speedup while maintaining visual fidelity, expressive motion, prompt responsiveness, and long-horizon identity consistency, demonstrating that our novel distillation and history replay make high-quality avatar generation practical for production.
% We presented \textbf{AptAvatar}, a two-step framework for fast and vivid
% long-form audio-driven avatar generation. By combining Endpoint-Anchored
% Distribution Distillation with Self-Generated History Replay, AptAvatar preserves
% the capacity and bidirectional temporal modeling of a 14B generator while
% reducing inference to only 2 NFEs. Experiments on 720p short- and long-form
% benchmarks show that AptAvatar achieves a 60$\times$ speedup while maintaining
% visual fidelity, expressive motion, prompt responsiveness, and long-horizon
% identity consistency. These results demonstrate that carefully designed
% distillation and history replay can make high-quality avatar generation practical
% for production-ready applications.
% ===== End sec/5_conclusion.tex =====
\bibliography{aaai2027}

% The AAAI-27 reproducibility checklist should be completed and uploaded separately
% in the OpenReview submission form unless the venue form explicitly instructs otherwise.
% The local checklist template is ReproducibilityChecklist.tex.

\end{document}